\documentclass[a4paper,USenglish,cleveref, autoref, thm-restate]{tgdk-v2021}

\usepackage{graphicx} 
\usepackage{paralist} 

\usepackage{xspace}

\newcommand{\ie}{\textit{i}.\textit{e}.,\xspace}
\newcommand{\eg}{\textit{e}.\textit{g}.,\xspace}

\title{
	Knowledge Graphs for the Life Sciences: Recent Developments, Challenges and Opportunities\footnote{Authors are listed in alphabetic order with authors' contributions presented at the end of the article. Jiaoyan Chen, Hang Dong, Janna Hastings, and Valentina Tamma are corresponding authors.}}

\titlerunning{Knowledge Graphs for Life Sciences}

\author{Jiaoyan Chen}{Department of Computer Science, University of Manchester, UK \and Department of Computer Science, University of Oxford, UK }{jiaoyan.chen@manchester.ac.uk}{https://orcid.org/0000-0003-4643-6750}{supported by the EPSRC project ConCur (EP/V050869/1).}

\author{Hang Dong}{Department of Computer Science, University of Oxford, UK}{hang.dong@cs.ox.ac.uk}{https://orcid.org/0000-0001-6828-6891}{supported by the EPSRC project ConCur (EP/V050869/1).}

\author{Janna Hastings}{Institute for Implementation Science in Health Care, University of Zurich, Switzerland \and School of Medicine, University of St. Gallen, Switzerland}{janna.hastings@uzh.ch}{https://orcid.org/0000-0002-3469-4923}{supported by the School of Medicine of the University of St. Gallen.}

\author{Ernesto Jim\'enez-Ruiz}{City, University of London, UK \and 
	SIRIUS, University of Oslo, Norway \and \url{https://www.city.ac.uk/about/people/academics/ernesto-jimenez-ruiz}}{ernesto.jimenez-ruiz@city.ac.uk}{https://orcid.org/0000-0002-9083-4599}{supported by the SIRIUS Centre for Scalable Data Access (Research Council of Norway, project 237889).}

\author{Vanessa L\'{o}pez}{IBM Research Europe, Dublin, Ireland}{vanlopez@ie.ibm.com}{https://orcid.org/0000-0002-0674-5324}{}

\author{Pierre Monnin}{Université Côte d’Azur, Inria, CNRS, I3S, France \and \url{https://pmonnin.github.io}}{pierre.monnin@inria.fr}{https://orcid.org/0000-0002-2017-8426}{}

\author{Catia Pesquita}{LASIGE, Faculdade de Ciências, Universidade de Lisboa, Portugal}{clpesquita@fc.ul.pt}{https://orcid.org/0000-0002-1847-9393}{funded by the FCT through LASIGE Research Unit (ref. UIDB/00408/2020 and ref. UIDP/00408/2020), and also partially supported project 41, HfPT: Health from Portugal, funded by the Portuguese Plano de Recuperação e Resiliência.
}

\author{Petr \v{S}koda}{Department of Software Engineering, Faculty of Mathematics and Physics, Charles University, Prague, Czechia}{petr.skoda@matfyz.cuni.cz}{https://orcid.org/0000-0002-2732-9370}{}

\author{Valentina Tamma}{Department of Computer Science, University of Liverpool, UK}{v.tamma@liverpool.ac.uk}{https://orcid.org/0000-0002-1320-610X}{}

\authorrunning{Chen, Dong, Hastings, Jim\'enez-Ruiz, L\'{o}pez, Monnin, Pesquita, \v{S}koda, Tamma}

\Copyright{Jiaoyan Chen, Hang Dong, Janna Hastings, Ernesto Jim\'enez-Ruiz, Vanessa L\'{o}pez, Pierre Monnin, Catia Pesquita, Petr \v{S}koda, and Valentina Tamma}



\category{Position} 

\relatedversion{} 

\acknowledgements{We would like to thank Uli Sattler (University of Manchester) for proposing the topic of this paper and Terry Payne (University of Liverpool) for the useful comments on a previous draft. We would also like to thank the TGDK editors in chief for organizing this inaugural issue.}

\nolinenumbers 

\Volume{1}
\Issue{1}
\ArticleNo{5}
\DateSubmission{Date of submission}
\DateAcceptance{Date of acceptance}
\DatePublished{Date of publishing}
\SectionAreaEditor{TGDK section area editor}

\begin{document}

	\maketitle
	\begin{abstract}
		The term \textit{life sciences} refers to the disciplines that study living organisms and life processes, and include chemistry, biology, medicine, and a range of other related disciplines. Research efforts in life sciences are heavily data-driven, as they produce and consume vast amounts of scientific data, much of which is intrinsically relational and graph-structured. 
		
		The volume of data and the complexity of scientific concepts and relations referred to therein promote the application of advanced knowledge-driven technologies for managing and interpreting data, with the ultimate aim to advance scientific discovery.
		
		In this survey and position paper, we discuss recent developments and advances in the use of graph-based technologies in life sciences and set out a vision for how these technologies will impact these fields into the future. We focus on three broad topics: the construction and management of Knowledge Graphs (KGs), the use of KGs and associated technologies in the discovery of new knowledge, and the use of KGs in artificial intelligence applications to support explanations (explainable AI). We select a few exemplary use cases for each topic, discuss the challenges and open research questions within these topics, and conclude with a perspective and outlook that summarizes the overarching challenges and their potential solutions as a guide for future research.
	\end{abstract}
	
	\section{Introduction}
	
	The term \textit{life sciences} refers to those disciplines that study living organisms and life processes, and include chemistry, biology, medicine, and a range of other related areas. Research efforts in life sciences are increasingly data-driven, as they produce and consume vast amounts of scientific data, much of which is intrinsically relational and graph-structured. 
	
	Much of this data is large-scale, complex, and presents many interrelationships and dependencies, thus being well suited to be represented in graph structures. For this reason, graph-based technologies are frequently used in the life sciences, and these disciplines have been drivers and early adopters of innovative  methods and associated technologies. 
	
\begin{figure*}[t!]
		\includegraphics[width=\textwidth]{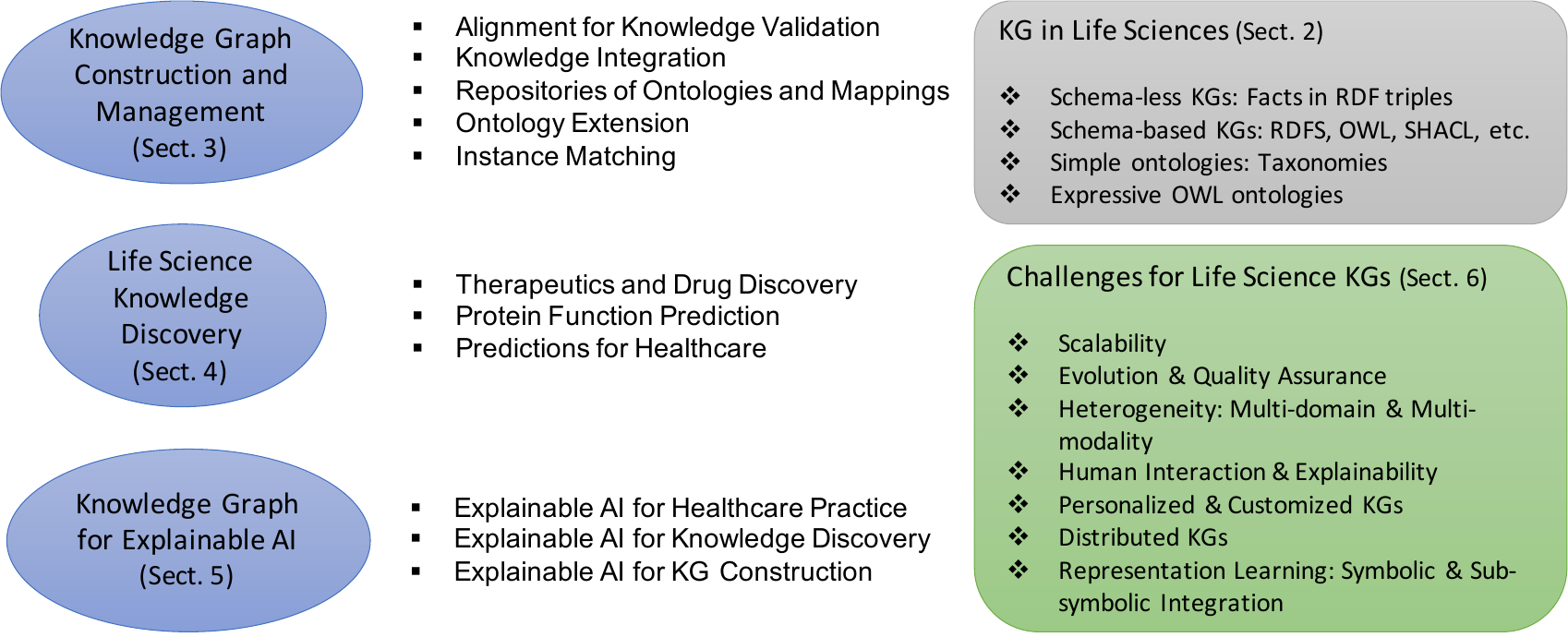}
		\caption{An overview illustration of definitions (upper right, in gray), topics (left column, in blue), use cases (middle), and challenges (bottom right, in green) for the research of KGs in the life sciences.}\label{life-sci-KG-illustration}
	\end{figure*}

	In this brief survey and position paper we discuss recent developments and advances in the use of graph-based technologies in life sciences, and set out a vision for how these technologies will impact these fields in future. 
	We illustrate the contribution in this paper in Figure \ref{life-sci-KG-illustration}. 
	
	We consider Knowledge Graphs (KGs) and their associated technologies to broadly include 
	\begin{inparaenum}[\it (i)]
		\item different forms of graph-based representations,
		\item the logical languages that assign explicit semantics to such representations, and their associated automated reasoning technologies, and 
		\item machine learning approaches that ingest data in graph-based representations and that process these graph-based representations to perform some task, \eg data analytics.
	\end{inparaenum}
	
	These different forms of graph-based representations can be further categorized based on the type of content represented. We therefore distinguish schema-less and schema-based Knowledge Graphs. More specifically, a typical KG contains either or both a schema part (terminologies or TBox\footnote{We introduce a list of key terms relevant to Knowledge Graphs and Life Sciences in Appendix A.}) and a data part (facts, assertions, or ABox). The formal semantics of KGs can be expressed with the OWL ontology language\footnote{Web Ontology Language: \url{https://www.w3.org/OWL/}}.
	
	In the remainder of this paper we will focus on three broad topic areas in which graph-based technologies have been used extensively, and we illustrate each area with some specific projects or use cases that guide our discussion and summary of the challenges that have been encountered. 
	
	\begin{itemize}
		\item The construction and management of KGs to represent life science knowledge; 
		\item The use of KGs and associated technologies in the discovery of new knowledge;
		\item The use of KGs in artificial intelligence applications to support explanations (eXplainable AI or XAI).
	\end{itemize}
	
	We then provide a summary of the general challenges across the topics, that include intrinsic characteristics of KGs (\eg scalability, evolution, heterogeneity) and their operational aspects in the real world (\eg human interaction, personalization, distributed setting, and representation learning). We present the challenges by means of use cases and the current research efforts that address them. 
	It is worth mentioning that while we aim to focus on the life sciences, many of the topics and challenges discussed in this work, especially those of KG construction and management in Section 3, are general and applicable to KGs in other domains such as finance, e-commerce, material, and urban management \cite{liu2023urbankg,deng2023construction}, etc.
	The KG-based problem modeling and solving approaches in life science knowledge discovery could be applicable for addressing many other use cases and problems in a broader domain of AI for scientific discovery \cite{wang2023scientific,hastings2023ai}.
	
	In the next section, we introduce several different categories of KGs as they have been used in  life sciences. Thereafter in Sections \ref{sec:KG_construc}-\ref{sec:KG_explainable_AI}, each of the above topics is described in a dedicated section together with a survey of recent advances. Finally, in Section~\ref{sec:Discussion} we synthesize the overarching challenges and trends into a perspective on the outlook for the future. 
	
	\section{Knowledge Graphs in the Life Sciences}\label{sec:KG_life_sci_def}
	
	KGs represent semantically-described real-world entities, typically through ontologies (vocabularies or schemas)~\cite{Hoehndorf-et-al2015:ontologies,dessimoz_primer_2017} and the data instantiating them, and thus provide descriptions of the entities  of interest and their interrelations, by means of links to ontology classes describing them, organized in a graph~\cite{sousa2020evolving}. 
	KGs have been widely adopted in the life sciences, as can be seen in the composition of the Linked Open Data Cloud\footnote{\url{http://cas.lod-cloud.net}}, where  life sciences represent one of the largest subdomains. A prominent example is the KG representing annotations regarding proteins by means of terms in the Gene Ontology describing different protein functions~\cite{ashburner2000gene}.
	
	Whilst KGs are becoming increasingly popular in different domains including the life sciences, there is no single accepted definition of KG~\cite{Ehrlinger2016TowardsAD}. A KG can be formally described as a directed, edge-labeled graph $\mathcal{G} = (V, E)$, where $V$ refers to the \emph{vertices} or \emph{nodes}, representing real-world entities of interest (\eg proteins, genes, compounds, cellular components, but also pathways, biological processes and molecular functions, to name a few) while $E$ refers to the edges in the graph, representing relationships or links between the entities in $V$ (\eg binds, associates, etc.). These may be represented as statements about entities in the form of RDF\footnote{Resource Description Framework: \url{https://www.w3.org/RDF/}} triples: (subject, predicate, object). 
	
	However, this formal definition only focuses on the components of KGs, but does not pose any constraint on what a KG should model or represent, and how. This is particularly true in life sciences, where the term \emph{Knowledge Graph} has been used to refer to diverse graph data structures, typically interconnected, but often isolated. 
	
	Many of the everyday tasks faced by researchers in this domain require the systematic processing and integration of data and knowledge from data sources that are characterized by heterogeneous syntaxes and structures, formats, entity notation, schemas and scope, \eg ranging from molecular mechanisms to phenotypes. Researchers in this area have been early adopters of Semantic Web and linked data approaches as a means to facilitate knowledge integration and processing to support tasks including semantic search, clinical decision support, enrichment analysis, data annotation and integration. However, a recent analysis of life science open data has identified several stand-alone data sources that exist in isolation, are not interlinked with other sources, and are schema-less (or use unpublished schemas), with limited reuse or mappings to other data sources~\cite{Kamdar2020AnEM}. Therefore, we can define a life sciences KG, following~\cite{NICHOLSON20201414}, as a data resource integrating one or more possibly curated sources of information into a graph whose nodes represent entities and edges represent relationships between two entities. This definition is consistent with other definitions found in the literature, \eg~\cite{10.3233/SW-160218}.
	
	These considerations underlie the reasons why KGs in life sciences can be of different types, and can be categorized across different dimensions. One of the most critical dimensions (in terms of support for complex queries and integration) is the categorization of KGs into schema-based and schema-less knowledge bases. In turn, the expressivity of the schema provides a further categorization criterion, depending on whether schemas are modeled as simple taxonomies (\eg the NCBI taxonomy \cite{Schoch2020ncbitaxonomy} included in the UMLS Metathesaurus \cite{DBLP:journals/nar/Bodenreider04}), RDFS\footnote{RDF Schema: \url{https://www.w3.org/TR/rdf-schema/}} vocabularies or (fully axiomatized) OWL ontologies. In particular, this paper refers to this broad definition of KGs, which we then divide into:
	\begin{itemize}
		\item Schema-less KGs composed of only relational facts in the form of RDF triples. Examples include the PharmaGKB dataset, an integrated online knowledge resource capturing how genetic variation contributes to variation in drug response~\cite{PharmGKBDataset2021}. Note that many semantic networks (defined in Appendix A) could be assigned to this category as their triples form a multi-relational graph.  
		
		\item Schema-based KGs composed of relational facts and their schema (meta information) in \eg RDFS, OWL, and constraint languages such as  SHACL\footnote{\url{https://www.w3.org/TR/shacl/}}. Examples include Wikidata with its property constraints, and DBpedia with its DBpedia ontology. Whilst Wikidata and DBpedia are general-purpose KGs, they also include large-scale life science knowledge.
		
		\item Simple ontologies representing taxonomies. Notable examples include the tree structure of the UMLS Semantic Network\footnote{\url{https://uts.nlm.nih.gov/uts/umls/semantic-network/root}} and the International Classification of Diseases, version 10 (ICD-10) \cite{icd10}.
		
		\item Expressive OWL ontologies, with complex axioms beyond simple taxonomies. OWL ontologies may be composed of a TBox and an ABox. Depending on the expressivity of the axioms modeled in the ontology, \ie the basic statements that an OWL ontology expresses, 
		OWL ontologies can fall into one of the previous categories: for instance, an OWL ontology with just an ABox can be seen as the case above of a KG composed of relational facts alone.
		In this final category we include fully axiomatized OWL ontologies, \eg with complex classes and property restrictions. Notable examples of these ontologies include SNOMED CT \cite{donnelly2006snomed}, the Gene Ontology \cite{ashburner2000gene,GO2023}, and the Food Ontology (FoodOn)\footnote{\url{http://foodon.org}}.
	\end{itemize}

	\section{Knowledge Graph Construction and Management}\label{sec:KG_construc}
	The adoption of KGs in the life sciences is motivated by the need for standardization of taxonomies and vocabularies to support the integration, exchange and analysis of data. More recently, richly annotated data is also being used in combination with machine learning methods for many applications, including helping to overcome issues related to the sparsity of data and helping to select promising candidates for reducing expensive and time-consuming physical experiments~\cite{DBLP:conf/semweb/HeCDJHH22}. 
	Graph-based machine learning approaches such as Graph Neural Networks have been applied to a number of life science tasks~\cite{Gaudelet-et-al2021}, including drug repurposing~\cite{morselli2021network} and predicting polypharmacy side effects~\cite{zitnik2018modelling}. 
	
	Given the diverse nature of the knowledge and tasks supported by KGs, the focus of state-of-the-art approaches has been the description of how individual KGs are developed within the specific domain~\cite{Yuan-et-al2019}, typically in terms of the specific approaches used for the development of the KG (\eg data extraction process, relation extraction and entity discovery), rather than on the overall development process. More recently, some efforts have focused on providing an overview of development approaches and pipelines for the construction of KGs in the life sciences, and beyond~\cite{NICHOLSON20201414, Tamasauskaitundefined-Groth2022}. The process of constructing a KG depends heavily on: 
	\begin{itemize}
		\item The type of data sources integrated and annotated by the KG, \eg CSV files, public and proprietary data sources, structured databases, full-text publications, etc.
		\item The granularity of the KG to be constructed, \eg schema-less KG, simple or expressive ontology.
		\item The usability expectations in downstream applications, \eg the ability to customize and manipulate the graph to support different use cases, or the ease of consumption as input to machine learning methods~\cite{geleta2021biological}. 
	\end{itemize}
	
	A recent systematic review~\cite{Tamasauskaitundefined-Groth2022} surveyed different KG development approaches to determine a general development framework. The review identified six main phases that  are common across different KG development approaches:
	\begin{itemize}
		\item[1)] Data source selection.
		\item[2)] Ontology construction.
		\item[3)] Knowledge extraction.
		\item[4)] Knowledge ingestion and validation.
		\item[5)] KG storage and inspection.
		\item[6)] KG maintenance and evolution.    
	\end{itemize}

	In the remainder of this section we will present the individual phases and the role they play in a KG development process by means of two use cases, where we illustrate the construction of KGs and discuss how these support knowledge integration and validation (Section~\ref{sec:Examples of KG construction}). We then present some recent technical developments in Section~\ref{sec:tasks-developments}, while Section~\ref{sec:kgc-challenges} discusses open challenges for the construction and management of KGs.
	
	\subsection{Knowledge Graph Construction Phases}\label{subsec:phases}
	This section provides more details on the phases involved in the KG construction process, with the aim of identifying recent trends, rather than providing an exhaustive literature survey. These phases are discussed in order of execution, however the \textit{ontology construction} phase can occur either together with the data source selection (if an ontology covering the domain of interest already exists or can be constructed through a set of given requirements) or as part of the \textit{knowledge ingestion and validation} phase, where an ontology is built semi-automatically from the available data or through modularization and alignment of existing ontologies.
	
	\subsubsection{Data source selection}
	This phase identifies the data sources that are to be integrated by the KG, which in turn affects the choice of knowledge extraction techniques. Generally, life science KGs ingest knowledge from structured, semi-structured and unstructured data sources. By \textit{structured} we refer to data modeled according to an existing structure, \eg data in tables or public or proprietary reference (relational) databases such as UniProt~\cite{UniProt} or ChEMBL~\cite{Gaulton-et-al2012}. Semi-structured data refer to, \eg XML documents~\cite{McCusker-et-al2020}, whereas unstructured data refer to data that do not conform to a given structure, \ie free-text sources, such as scientific publications from PubMed\footnote{\url{https://pubmed.ncbi.nlm.nih.gov}}. 
	Data ingested from manually curated databases~\cite{NICHOLSON20201414} and semi-structured sources constitute the foundation of a KG~\cite{geleta2021biological}, generally defining the entities and some of the relations in the KG. This data is then further enriched by performing text mining on large-scale free text sources, in order to extract relationships, which is the objective of the \textit{knowledge extraction} phase.
	
	\subsubsection{Ontology construction}
	The aim of this phase is to define a common, consensus-based, controlled vocabulary to describe the data in an \emph{ontology}~\cite{rector2019beyond}. The existence of a common structure, or schema, supports querying, integration and reasoning tasks over the KG. 
	
	Traditional ontology engineering approaches are divided into top-down or bottom-up. Top-down approaches are based on more or less formal ontology engineering methodologies~\cite{fernandez1997methontology,Kendall_McGuinness_2019OE,Noy-McGuinness01} or common practices~\cite{10.7551/mitpress/9780262527811.001.0001} to build ontologies from a description of the domain elicited from domain experts \cite{neuhaus2022ontology}, and/or by reusing or extending existing ontologies \cite{DBLP:conf/esws/Jimenez-RuizGSSL08}. Ontology engineering methodologies define the ontology development process in terms of requirement analysis, entity and property definitions, ontology reuse, validation and population. In contrast, bottom-up approaches utilize semi-automatic data driven techniques, \eg ontology learning from text~\cite{liu2011natural}, and can be used to refine and validate an ontology. These approaches are discussed in more detail when presenting the \textit{knowledge ingestion and validation} phase. 
	
	Whilst general purpose ontology engineering methodologies have evolved to be used in the development of KGs~\cite{POVEDAVILLALON2022104755}, a considerable number of ontologies in the life science domain have been built as part of the Open Biological and Biomedical Ontologies (OBO) Foundry effort,\footnote{\url{https://obofoundry.org}} which defines a set of development principles for biological and biomedical ontologies and provides a suite of high-quality, interoperable, free and open source tools that support ontology development~\cite{10.1093/database/baac087}.

	\subsubsection{Knowledge extraction}
	Knowledge extraction refers to the identification of entities and their relations from the data sources, which is a crucial step in the development of a KG~\cite{Tamasauskaitundefined-Groth2022}. \textit{Entity extraction} identifies entities from the various data sources selected using Natural Language Processing (NLP) approaches and text mining techniques to analyze and extract relevant information from large text corpora~\cite{wang2018comparative, leser2005makes, huang2016community}. Named entity recognition (NER) supports the identification of named entities in text, such as drug names, diseases, or chemical compounds, and their classification according to pre-defined entity types~\cite{nadeau2007survey}.  NER approaches in the life sciences are typically based on labor intensive tasks such as the definition of generic (\eg orthographic, morphological, or dictionary-based) and  specific rules that are typically defined by experts, and are not easily applicable to other corpora~\cite{zhu2018gram}. There are a number of issues hindering these approaches:
	\begin{inparaenum}[a)]
		\item the pace of scientific discovery and the identification of new entities; 
		\item the large number of synonyms and term variations associated with an entity; and
		\item  entity identifiers that are composed of a mixture of letters, symbols and punctuation, often in large sentences~\cite{leaman2015challenges}.
	\end{inparaenum}
	More recent approaches have proposed the use of supervised machine learning methods (\eg conditional random fields, or Support Vector Machines, SVMs, neural networks, and neural language models in particular)~\cite{makino2002tuning,jurafsky2023slpbook,dong2023outKB} either in isolation, or combined in hybrid approaches to improve accuracy~\cite{rocktaschel2012chemspot}. 
	
	Entity recognition generates entities that are isolated and not linked~\cite{Tamasauskaitundefined-Groth2022}. The goal of \textit{Relation extraction} is to discover relationships of interest between a pair of entities, thus describing their interaction. 
	Relation extraction is a necessary step for entities defined in semi-structured or unstructured sources, whereas structured data sources are characterized by explicitly identifiable relationships. Typical approaches for relation extraction include rule-based~\cite{hunter2008wab, ravikumar2014towards,Ravikumar2017BELMinerAA}, supervised~\cite{DBLP:journals/biodb/LiSJSWLDMWL16, Fundel2007DataAT} and unsupervised approaches~\cite{ea9bc3d4f3de49a0a93324cdb8c5718e, NICHOLSON20201414}. Rule-based relation extraction identifies keywords (based on existing ontologies or expert defined dictionaries) and grammatical patterns to discover relations between entities. Supervised relationship extraction methods utilize publicly available pre-labelled datasets (\eg BioInfer~\cite{pyysalo2007bioinfer} or BioCreative II~\cite{krallinger2008overview}) to construct generalized patterns that separate positive examples (sentences implying the existence of a relationship) from negative ones. Supervised approaches include SVMs, Recurrent Neural Networks (RNNs) and Convolutional Neural Networks (CNNs)~\cite{bach2007review, NICHOLSON20201414}. 
	Unsupervised relation extraction methods~\cite{10.1162/tacl_a_00095} have emerged to address the lack of scalability of supervised relation extraction methods, due to the high cost of human annotation. Unsupervised methods involve some form of clustering or statistical computation to detect the co-occurrence of two entities in the same text~\cite{NICHOLSON20201414}.
	
	More recently, end-to-end approaches (End-to-End Relation Extraction -- RE) have been used to tackle both tasks simultaneously. In this scenario, a model is trained simultaneously on both the NER and Relation Extraction objectives~\cite{huguet-cabot-navigli-2021-rebel-relation}. Furthermore, rule-based approaches can be combined with relation classification using specialized pre-trained language models adapted for life science domains, \eg BioBERT \cite{10.1093/bioinformatics/btz682}, SapBERT \cite{liu-etal-2021-self}, and RoBERTa-PM \cite{lewis-etal-2020-pretrained}, to name a few. There is also a recent trend to probe and prompt pre-trained language models to extract relations (\eg disease-to-disease, disease-to-symptoms) \cite{yao2022extracting,sung-etal-2021-language}. 
	
	\subsubsection{Knowledge ingestion and validation}
	The aim of this phase is to ingest the entities and relationships extracted in a previous phase, which models knowledge from different sources. These entities and relations can be incomplete, ambiguous or redundant, and need to be appropriately aligned and integrated, and finally annotated according to the ontology constructed in phase 2. 
	
	Knowledge integration or fusion can critically improve the quality of data by performing \textit{entity resolution}, \ie the detection of different descriptions of the same real-world entity (also called entity matching, deduplication, entity linkage or entity canonicalization), prior to ingesting them in the KG. 
	This reconciliation step is particularly crucial in the life sciences, where duplication can be caused by data modeled using different vocabularies or ontologies, or when data is extracted from literature sources that are rapidly changing. The severity of the ambiguity depends on the number of ontologies available for the domain. For instance, the number of gene vocabularies is far smaller than the number of disease vocabularies that could be present in the ingested datasets. Linking these entities requires costly alignment processing; in particular the alignment of disease entities is especially problematic given the number of  different coding systems, whose conversion is often not trivial~\cite{geleta2021biological}. We further explore this issue in two of the use cases presented in Section~\ref{sec:tasks-developments}, where we explore the problem of aligning vocabularies and ontologies through the use of mapping repositories and instance matching in automated clinical~coding. 
	
	Entities are assigned unique identifiers (URI or IRI) that support the definition of bespoke namespaces, and support integration by reusing identifiers in related namespaces. 
	Entity resolution is based on clustering similar entities together in a \textit{block}, where similarity measures are used to detect duplicates~\cite{Tamasauskaitundefined-Groth2022}. Typical methods include sorted neighborhoods and traditional blocking; and machine learning methods are commonly used for similarity computation, \eg feature vector computation~\cite{10.5555/3351851}.
	
	This phase may also include the bottom-up construction of the ontology for those applications where a top-down approach is not feasible. Bottom-up approaches extract the relevant knowledge first, and then they construct the data schema / ontology based on the extracted data, typically using (semi-)automated methods, based on machine learning.  Ontologies define the structure of the knowledge graph, which supports querying and data analytics. In bottom-up ontology development the structure of the knowledge graph is determined based on the extracted knowledge, thus providing a structure for this knowledge~\cite{kg-book}.
	
	Often the construction of ontologies (either bottom-up or top-down) relies on the ability to correctly align and reuse entities defined across different domains and KGs. Furthermore, reuse of (or conformance to) existing upper level ontologies, \eg BFO (Basic Formal Ontology)~\cite{10.7551/mitpress/9780262527811.001.0001}
	provides the basis for the consistent and unambiguous formal definition of entities and relations that prevents errors in coding and annotation. The alignment of ontologies in life sciences and other domains is an active area of research, and we provide an overview of recent technical developments and challenges in Section~\ref{sec:tasks-developments}.
	
	Whilst bottom-up approaches, especially those based on alignment, are becoming more viable, especially given the support of language models, such as BERT~\cite{he2023deeponto}, their performance is not always adequate for the task, as discussed in the second challenge in Section~\ref{sec:kgc-challenges}.
	
	Knowledge enrichment and completion improve the KG quality by performing reasoning (KG materialization), inference~\cite{guimaraes_et_al:OASIcs.AIB.2022.2} and optimization. Reasoning and inference support the assertion of new relations based either on logical reasoning (\eg \cite{DBLP:conf/semweb/NenovPMHWB15,DBLP:journals/pvldb/TsamouraCMU21}) or machine learning techniques (\eg statistical relational learning or through embedding based link predictors for new concepts~\cite{dong2023oet,dong2023outKB,heist2023nastylinker, Iurshina2022nilk} and node classifiers, also called KG refinement~\cite{10.3233/SW-160218}). The extent and type of logical inferences depends on the expressivity of the ontology built in phase 2, or in a bottom-up fashion in this phase, together with any associated mappings. Description Logic formalisms, such as OWL, use logic-based reasoning for detecting and correcting incorrect assertions and ontology alignments~\cite{chen2023assertion}.

	\subsubsection{KG storage and inspection}
	KGs need to be accessible to support a variety of different tasks, beyond the mere integration of different knowledge sources, and thus KG storage management~\cite{Tamasauskaitundefined-Groth2022,qi2023prekar,wang2020knowledge} is an active area of research. Current KG storage mechanisms are divided into relation based stores (\eg \cite{abadi2009sw}) and native graph stores (\eg~\cite{zou2014gstore}). Relational KG stores, either based on relational databases or through NOSQL databases and / or triple stores such as Jena TDB\footnote{\url{https://jena.apache.org/documentation/tdb/index.html}}, have reached a considerable level of maturity and have been optimized in order to avoid common problems, \eg a large number of null values in columns or optimized query performance~\cite{qi2023prekar}. Graph databases store nodes, edges and properties of graphs natively, and support query and graph mining tasks. Examples of state of the art implementations include Neo4J\footnote{\url{https://neo4j.com}}, GraphDB\footnote{\url{https://graphdb.ontotext.com}}, and RDFox\footnote{\url{https://www.oxfordsemantic.tech/product}}. The evolution of the performance of these systems has been the object of systematic studies~\cite{10.1145/3604932}, whereas~\cite{10.1093/database/baab026} explicitly focuses on biomedical use cases.
	
	Storage management has implications on the ways KGs support expressive queries for nodes and edges and visualization, to support data analysis, navigation and discovery of related knowledge~\cite{10.5555/3351851,Sun2016VisualizationFK}. Graph databases often provide built-in tools for visualization, \eg Neo4J, whereas different Javascript libraries (\eg SigmaJS\footnote{\url{https://github.com/jacomyal/sigma.js}}) are available for developing visualization front ends. Support for complex queries is also either built in a graph database or a triple store by supporting the SPARQL query language~\cite{2013sparql,zou2014gstore}, or proprietary query languages such as Cypher~\cite{10.1145/3183713.3190657}, supported by Neo4J.
	
	\subsubsection{Knowledge maintenance and evolution}
	
	Given the rapid scientific development in the life sciences, and the consequent continuous update of ontologies for this domain, artifacts annotated with these ontologies can become outdated very quickly, and require some form of update (also called ontology extension). These update mechanisms need to be automated to ensure that they scale to the size of KGs. Automatic update approaches are based on the periodical detection and extraction of new knowledge that is then mapped to existing entities and relations in~the~KG~\cite{Wu-et-al2018}. 
	
	Update mechanisms are typically based on the detection of \textit{changes}~\cite{motik2002oewoe} that can affect an ontology, \eg addition, removal or modification of meta-entities (\ie entities, relations and their definitions). These changes include renaming concepts and properties, setting domain and range restrictions, or setting a subsumption relation.
	To date, the most comprehensive account of ontology change is given in \cite{FMKPA08}, where change is described for different sub-fields, \eg ontology alignment, matching and mapping, morphisms, articulation, translation, evolution, debugging, versioning, integration and merging; each with different requirements and implications. The study \cite{pernisch2021bio-onto-evol-mater} further investigates the impact of biomedical ontology evolution on materialization. 
	
	Currently available tools and methodologies use (semi)-automated methods to perform many of the operations that trigger a change in an ontology and the consequent creation of a new version~\cite{GRO2016333,he2023deeponto}. Different ontology management platforms and portals mandate different principles and frameworks for handling ontology versioning (\eg OBO foundry\footnote{\url{http://www.obofoundry.org/principles/fp-004-versioning.html}} or BioPortal\footnote{\url{https://bioportal.bioontology.org}}), but these are typically implemented by ontology developers with limited tool support. Section~\ref{sec:tasks-developments} presents an example of automated ontology extension that relies on machine learning to cope with the scale of data.

	\subsection{Examples of Life Science KG Construction}~\label{sec:Examples of KG construction}
	In this section we provide two examples of life science KGs that illustrate in practice the phases composing the generic KG construction process discussed in Section~\ref{sec:KG_construc}; namely a KG for Pharmacogenomics, PGxLOD \cite{monninLHRTJNC19}, and one for Ecotoxicological Analysis, TERA \cite{myklebust2019knowledge,myklebust2022prediction}.
	
	\medskip
	\noindent\textbf{Alignment for Knowledge Validation: An Example of Pharmacogenomics.}
	As mentioned in Section~\ref{sec:KG_construc}, the task of aligning knowledge in KGs supports several downstream applications and domains. 
	For instance, pharmacogenomics studies the influence of genetic factors on drug response phenotypes  (\eg expected effect, side effect). 
	Hence, pharmacogenomics is of interest for personalized medicine. 
	The atomic knowledge unit in pharmacogenomics is a ternary relationship between a drug, a genetic factor, and a phenotype.
	Such a relationship states that a patient being treated with the specified drug while having the specified genetic factor may experience the described phenotype.
	Semantic Web and KG technologies have been employed in this application domain, for example by building ontologies in which patients and pharmacogenomic knowledge are represented, and then using deductive reasoning mechanisms to conditionally recommend genetic testing before drug prescription~\cite{samwaldGBFAD15}. 
	However, the knowledge relevant to pharmacogenomics is scattered across several sources including reference databases such as PharmGKB, and the biomedical literature.
	Additionally, this knowledge may lack sufficient validation to be implemented in clinical practice.
	For example, some relationships may have only been observed in smaller cohorts of patients or in non-replicated studies. 
	Hence, there is a need to align different sources of pharmacogenomic knowledge to detect additional evidence validating (or moderating) a knowledge unit.
	To this aim, the PGxLOD KG was proposed~\cite{monninLHRTJNC19}. 
	Automatic knowledge extraction approaches were applied on semi-structured and unstructured data from PharmGKB and the biomedical literature to represent their knowledge in the KG. 
	Then, matching approaches were developed to align knowledge units from various sources~\cite{monninCNC20,monninRNC22}.
	The resulting alignments outlined some agreements between PharmGKB and the biomedical literature, which was expected since PharmGKB is manually completed by experts after reviewing the literature.
	Interestingly, this automatic knowledge extraction pipeline could guide the manual review process by automatically pointing out studies confirming or mentioning a pharmacogenomic knowledge unit.
	
	\medskip
	\noindent\textbf{Knowledge Integration: An Example of Ecotoxicological Analysis.}
	In ecotoxicological analysis, data and knowledge from different domains such as chemistry and biology are often needed. These are usually located in different sources such as spreadsheets or CSV files for local experimental results, open databases for public research results, and ontologies for domain knowledge.
	Thus knowledge integration becomes a critical and fundamental challenge before real analysis can be conducted.
	In the study by Myklebust \textit{et al.} \cite{myklebust2019knowledge,myklebust2022prediction}, which aims to predict adverse biological effects of chemicals on species, a toxicological effect and risk assessment KG named TERA was constructed for knowledge integration.
	TERA includes three sub-KGs: \textit{(i)} the Chemical sub-KG, which is constructed by integrating the vocabulary MeSH (Medical Subject Headings) with selective knowledge from two chemical databases PubChem and ChEMBL utilizing the chemical mappings in Wikidata; \textit{(ii)} the Taxonomy sub-KG, which is constructed by integrating EOL (Environment Ontology for Livestock) and the NCBITaxon ontology utilizing NIBI-EOL mappings in Wikidata; and \textit{(iii)} the ECOTOX sub-KG, which is composed of RDF triples transformed from experimental risk results and is aligned with the other two sub-KGs by the ontology alignment system LogMap~\cite{jimenez2011logmap} and the chemical mappings in Wikidata. Another example of knowledge integration is for drug repurposing, where the KG Hetionet\footnote{\url{https://github.com/hetio/hetionet}} is created by integrating 29 public resources, including biomedical KGs and other types of data \cite{himmelstein2017kg-integration-drug}.
	
	\subsection{What has been done: recent technical developments}~\label{sec:tasks-developments}
	Given the many existing ontologies in life sciences, \eg ontologies available in the OBO Foundry collection or in  BioPortal~\cite{Noy2009bioportal}, KG construction usually involves the reuse, alignment, and enrichment of state-of-the-art ontologies. The existing ontologies in life sciences need to be updated given the new discoveries in the field. This is broadly a key issue in the management, maintenance, and evolution of ontologies. We select a few promising use cases below to highlight some recent developments that support the KG construction in the life sciences.
	
	\medskip
	\noindent
	\textbf{Repositories of Ontologies and Mappings.}
	Ontologies and their mappings play a central role in semantically enabled products and services consumed by life science companies, academic institutions and universities, as highlighted by the Pistoia Alliance ontology mapping project~\cite{drudis2019}.\footnote{\url{https://www.pistoiaalliance.org/projects/current-projects/ontologies-mapping/}} Ontology mappings are essential in knowledge graph construction tasks to bridge the knowledge provided by different ontologies and expand their coverage. 
	Ontology mappings can also play a key role
	when identifying the right ontologies to be reused as they will enable the retrieval of the relevant (overlapping) ontologies for the domain of interest. For this reason, a number of notable efforts in life sciences have created large repositories of ontologies and mappings to serve the research within the community. Prominent examples include the UMLS Metathesaurus~\cite{DBLP:journals/nar/Bodenreider04}, BioPortal~\cite{Noy2009bioportal,bioportal2013}, MONDO \cite{DBLP:conf/icbo/VasilevskyEMHHR20}, and the EBI services: OLS~\cite{DBLP:conf/edbt/VrousgouBPJ16}, OxO~\cite{DBLP:conf/icbo/JuppLSVBP17} and the RDF platform~\cite{DBLP:journals/bioinformatics/JuppMBBDGGGLRWMNPBJ14}.
	The UMLS Metathesaurus is a comprehensive effort for integrating biomedical ontologies through mappings. In its 2023AA version, it integrates more than two hundred vocabularies, with more than 3 million unique concepts and more than 15 million concept names. BioPortal is a repository containing more than 1,000 biomedical ontologies and more than 79 million lexically computed mappings among them (as of July 13, 2023). The Mondo Disease Ontology (MONDO) is a manually curated effort to harmonize and integrate disease conceptualizations and definitions across state-of-the-art ontologies (\eg HPO \cite{kohler2021human}, DO \cite{schriml2022human}, ICD, SNOMED CT, etc.). The services provided by the European Bioinformatics Institute (EBI) also deserve a special mention. The Ontology Lookup Service (OLS) has become a reference to explore the latest versions of more than two hundred ontologies via its graphical interface or programmatically via its API. OxO is a repository of ontology mappings and cross-references extracted from the OLS and UMLS. OxO allows users to visually traverse the graph of mappings to identify additional potential mappings beyond direct ones (\ie multi-hop mappings). Finally, the EBI RDF platform provides a unified KG with all the RDF resources at the EBI. 
	Complementary to the efforts from the life sciences, the Semantic Web has also contributed to the systematic evaluation of mappings in public repositories (\eg \cite{DBLP:journals/biomedsem/Jimenez-RuizGHL11,DBLP:conf/semweb/FariaJPSC14}) and mappings produced by automated ontology mapping systems (\eg the Ontology Alignment Evaluation Initiative (OAEI) \cite{DBLP:conf/semweb/PourABCC0FFFH0022}). Automatically generated mappings of high quality have the potential to be integrated within the aforementioned repositories and hence, the OAEI has always had a special focus on life science test cases with evaluation tracks like Anatomy \cite{DBLP:journals/biomedsem/DragisicILL17}, LargeBio \cite{DBLP:conf/dlog/Jimenez-RuizMGH13}, Phenotype \cite{DBLP:journals/biomedsem/HarrowJSRWMAKMW17} and the newly created track BioML \cite{DBLP:conf/semweb/HeCDJHH22}. 
	The Simple Standard for Sharing Ontological Mappings (SSSOM) \cite{DBLP:journals/biodb/MatentzogluBBBB22} represents a joint effort between the life sciences and Semantic Web communities to facilitate the exchange of mappings across different parties and repositories, while keeping the provenance and other relevant characteristics of the mappings. 
	
	\medskip
	\noindent\textbf{Ontology Extension.} 
	Ontology extension in life sciences aims to connect new concepts and their relations to an ontology from updated sources, \eg scientific papers in PubMed and chemical information in PubChem\footnote{\url{https://pubchem.ncbi.nlm.nih.gov/}}. Manual ontology extension, while essential for the development of gold standard resources, is not scalable to the full scope of large domains due to its high cost and low efficiency, and sometimes is even unfeasible as human beings may not be able to review the quantities of new information at the rate they become available. Thus machine-learning-based, automated methods are needed. One recent example is the use of deep learning, specifically a Transformer-based model, to categorize new chemical entities within the ChEBI ontology\footnote{\url{https://www.ebi.ac.uk/chebi/}} \cite{glauer_interpretable_2022}. In addition, recent studies have explored enriching SNOMED~CT by mining new concepts from texts \cite{dong2023outKB} and placing them into the ontology \cite{Liu2020placement,dong2023oet}. A new concept can be identified by NIL entity linking, \ie exploring unlinkable mentions, usually through setting a ``linkable'' score threshold or through classification \cite{dong2023outKB}. Resolution and disambiguation of NIL mentions with clustering can help to represent NIL entities \cite{heist2023nastylinker,kassner2022edin}. For concept placement, similar to the aforementioned CHEBI ontology extension \cite{glauer_interpretable_2022}, machine learning, especially in the form of Transformer-based deep learning, has been applied to predict subsumption relations between a new concept and the existing concepts. Complex concepts in OWL ontologies that contain logical operators (\eg existential quantifier and  conjunction in SNOMED CT) can be supported in subsumption prediction \cite{chen2023bertsub} and new concept placement \cite{dong2023oet}. Another group of studies use post-coordination or formalising a new term with existing concepts and attributes \cite{castell2023postcoordination,kate2020}, which is similar to composing subsumption axioms with complex concepts. The methods include using lexical features \cite{kate2020}, word embeddings and KG embeddings \cite{castell2023postcoordination}. Pre-trained and Large Language Models, through fine-tuning, zero-shot and few-shot prompting have the potential to support the mining \cite{dong2023outKB} and placement of new concepts (\eg by subsumption prediction \cite{chen2023bertsub,he2023language}).
	
	\medskip
	\noindent\textbf{Instance Matching: Automated Clinical Coding.} A main source for patients' KG construction is Electronic Health Records (EHR). Using medical ontologies as backbones, it is possible to add a layer of data by instance matching (or patient matching) through \textit{Clinical Coding}. Clinical coding is the task of transforming medical information in EHR into structured codes described in medical ontologies \cite{dong2022automated}, \eg ICD and SNOMED CT. Recent approaches mainly formulate the problem as a multi-label classification problem. Various neural network architectures have been proposed and knowledge plays a key role to enhance the neural architectures \cite{dong2022automated,ji2022unified}. Pre-trained language models, \eg BERT \cite{devlin-etal-2019-bert}, have been applied to clinical coding and gradually achieved better results with adapted modeling methods and more advanced language models, \eg PLM-ICD \cite{huang-etal-2022-plm} with RoBERTa-PM \cite{lewis-etal-2020-pretrained}, according to studies \cite{dong2022automated,Edin2023medCoding,ji2021clinicalCodingwithBERT}. Other studies formulate the task as a Named Entity Recognition and Linking (NER+L) problem, by extraction of concepts and linking them with the ontologies \cite{dong2022automated}. Overall, the recent progress in clinical coding, along with the advent of Large Language Models (LLMs) suggests a trend in this area for patients' KG construction from EHR. However, there is still room for improvement in knowledge integration to better address explainability (see Section \ref{sec:KG_explainable_AI} for more details) and in zero-shot learning problems, \ie for classifying into rare codes or concepts \cite{dong2022automated,Edin2023medCoding, ji2022unified}. There are also further recent examples of instance matching with EHR data, including the works \cite{carvalho2023knowledge,theodoropoulos-2023}.
	
	\subsection{What are the challenges?}\label{sec:kgc-challenges}
	
	KG construction and management often play a fundamental role in supporting life sciences with computation.
	There are still quite a few technical challenges, and many of the current tools and algorithms can be improved by modern machine learning and AI techniques. 
	Here we present some critical and fundamental technical challenges.
	\begin{itemize}
		\item \textbf{How to construct a customized KG?} For a specific application, we often need to extract relevant data and knowledge from multiple sources, and at the same time integrate extracted knowledge from different sources. Considering a case study of personal health assistance, a customized KG with knowledge of at least exercise (sports), food, disease and medicine are required, while fine-grained knowledge of these aspects will lie in different domain KGs. 
		The key challenge for integrating different ontology modules lies in estimating the semantic similarity and discovering the equivalence of two knowledge elements with their contexts considered, as well as the subsequent refinement like KG completion and knowledge representation canonicalization.
		Adequate tool support to minimize manual curation but enabling the user involvement when required is also paramount (\eg \cite{DBLP:journals/ker/LiDFIJLP19}). 
		\item \textbf{How to ensure adequate performance using machine learning based approaches for automated KG construction?} At the TBox level, the state-of-the-art alignment between classes (especially for subsumption relations) seems to not yet be achieving good enough performance, as reflected in recent biomedical ontology alignment benchmarking \cite{DBLP:conf/semweb/HeCDJHH22}. At the ABox level, predicting missing facts for practical KG construction expects high precision (\eg beyond 90\% or 95\%) but only a few relations can be populated with a precision above 80\% using prompt learning with BERT as evaluated in \cite{veseli2023evaluating}. This is also the case to associate patients' EHR (as a part of ABox) with clinical codes or concepts in medical ontologies, where a micro $F_1$ score is below 60\% \cite{dong2022automated}. Learning subsymbolic representations (see defined in Appendix A) of KG and data sources may help address the challenge. 
		Transformer-based language models have achieved great performance in recent years. Among them, pre-trained language models such as BERT have been applied for KG construction with a promising performance achieved (see \eg the package DeepOnto \cite{he2023deeponto}), while the more recent and more powerful generative language models like GPT series \cite{brown2020language} have not been well applied at the time of writing, especially in the life science domain.
		
		\item \textbf{How to ensure reliable semi-automated deep learning-based KG construction with human interaction?} Many tasks in the KG life cycle unavoidably rely on human experts to achieve consensus on reliable knowledge; on the other hand, as the automated KG construction process is growing opaque with deep learning methods, it is important to ensure trustworthiness and reliability \cite{zhang2023expHumanKGC}. Apart from enhancing performance metrics with novel methods, results with certain explainability are needed, for example, highlighting key parts in the data input when they are used as sources for KG construction. We discuss other aspects of explainability with KG, on life science knowledge discovery and healthcare decision making, in Section \ref{sec:KG_explainable_AI}. Human-in-the-loop learning design for explainable KG construction may ensure the use of experts' knowledge for the task across the KG life cycle, which still remains a challenge for future research \cite{zhang2023expHumanKGC}.
	\end{itemize}

	\section{Life Science Knowledge Discovery}\label{sec:Knowledge-Discovery}
	
	Research into AI technologies -- including machine learning and KG-based reasoning -- to accelerate the pace of scientific discovery is an emerging and rapidly developing field. The challenge lies in assisting scientists to uncover new knowledge and solutions, such as discovering novel therapeutic opportunities, identifying candidate molecular drugs to treat complex diseases or alternatively new uses for existing drugs, and supporting more personalized predictions.
	
	Knowledge Graphs are powerful tools for representing complex biomedical knowledge, including molecular interactions, signaling pathways, disease co-morbidities, and more. Overviews of graph representation learning in biomedicine for healthcare applications and polypharmacy tasks are presented in \cite{li2022graph} and \cite{gema2023knowledge} respectively. In graph representation learning, the graph's topology is leveraged to create compact vector embeddings. Through nonlinear transformations, high-dimensional information about a node’s graph neighborhood is distilled into low-dimensional vectors, where similar nodes are embedded close together in the vectorial space. Embeddings have been shown to be valuable for handling numerous relations in a KG while efficiently exploiting relation sparsity using vector computations. These optimized representations are subsequently used to train downstream models for various tasks, such as predicting property values of specific nodes (\eg protein function), predicting links between nodes (\eg binding affinity between molecules and protein targets), or performing classification tasks (\eg predicting the toxicity profile of a candidate drug, or risk of readmission for a patient). 
	
	It is worth mentioning that among the existing works for life science knowledge discovery, different kinds of KGs have been exploited. 
	The schema-less KG can be used to model different kinds of interaction between instances such as proteins and drugs; the taxonomy alike simple ontology is often used to represent concepts and their hierarchy such as protein functions defined in the gene ontology, chemical compounds, species, and diseases; expressive OWL ontologies and schema-based KGs can be used to model complex logical relationships between concepts, besides simple interaction between instances. Such diverse knowledge representation capabilities make KGs more flexible in modeling the input data and prediction targets of different knowledge discovery tasks, than graphs and tabular data that are widely used in previous pure machine learning-based methods.
	
	In the following, we present some typical use cases, where machine learning techniques (including graph representation learning and language models) are applied over KGs built from diverse sources and domain ontologies, to facilitate life science discovery.

	\subsection{What has been done: use cases and their recent developments}\label{sec:KnowDisc-dev}
	
	\noindent\textbf{Therapeutics and Drug Discovery: Learning a representation using multi-modal and heterogeneous knowledge.} Drug discovery entails exploring an extremely large space of potential drug candidates. AI can help to accelerate this process by narrowing down the most promising  candidates before expensive experimentation. The key to leveraging predictive and generative models for candidate solution generation lies in learning an effective multi-modal representation of protein targets, molecules and diseases among others. 
	Recent research has focused on applying language models over large databases of proteins or molecules for self-supervised representation learning, such as ESM \cite{rivesetal_2021} and ProteinBERT \cite{proteinbert} for protein sequences, or Molformer for the molecule simplified molecular-input line-entry system (SMILES) \cite{ross2022molformer}. These models have exhibited remarkable success in tasks such as predicting protein interactions, binding affinity between drugs and targets, and protein functions and structures. However, these existing pre-trained sequence-based models often neglect to incorporate background knowledge from diverse sources, for example, biological structural knowledge. 
	
	Nonetheless, recent research indicates that incorporating existing expressive factual knowledge can improve results in downstream machine learning tasks. To enhance Protein Language Models (PLM), approaches such as OntoProtein \cite{zhang2022ontoprotein} and KeAP \cite{zhou2023protein}  use a KG of protein sequences augmented with textual annotations from the Gene Ontology (GO). OntoProtein was the first to inject gene ontology descriptions into a PLM for sequences to predict protein interactions, function and contact prediction. OntoProtein proposes to reconstruct masked amino acids while minimizing the embedding distance between the contextual representation of proteins and associated knowledge terms. Similarly, ProtST \cite{xu2023protst} uses a dataset of protein sequences augmented with textual property descriptions from biomedical texts and jointly trains a PLM with a biomedical language model. 
	
	Knowledge Graphs are suitable data models for expressing heterogeneous knowledge and facilitating end-to-end learning \cite{Wilcke-2017}. An entity in a KG can have multiple attributes with different modalities - where each modality provides extra information about the entity - as well as relations to and from entities in other sources. Graph Neural Networks (GNN) have been used to capture inter-dependencies and diverse types of interactions between heterogeneous entity types and multimodal attributes in KGs \cite{lam-otter}. They achieve this by iteratively aggregating information from neighboring nodes (through a process called message passing) and employing scoring functions to optimize the learned embeddings for downstream tasks. Otter-Knowledge \cite{lam-otter} incorporates a heterogeneous KG (schema-based, containing concepts and their attributes) from diverse sources and modalities, \ie each node has a particular mode that qualifies its type (text, image, protein sequence, molecule, etc.) and initial embeddings for each node are computed based on their modality. A GNN is then used to enrich protein and molecule representations and train a model to produce final node embeddings. The model is able to produce representations for entities that were not seen during training and achieve state-of-the-art results in the Therapeutic Data Commons (TDC) benchmarks \cite{Huang2022} for drug-target binding affinity prediction.
	TxGNN \cite{huang2023zeroshot} uses a GNN pre-trained on a large heterogeneous, multi-relational KG of diseases and therapeutic candidates constructed from various knowledge bases. TxGNN obtains a signature vector for each disease based on its neighboring proteins, exposure and other biomedical entities to compute a disease similarity and predict drug indication/contraindication for poorly characterized diseases.
	
	\medskip
	\noindent\textbf{Protein Function Prediction with the Gene Ontology.} Conducting physical experiments for identifying protein functions is time and resource consuming.
	With the development of machine learning, protein function prediction (which is the task of predicting a given protein with multiple and potentially hierarchical classes -- functions -- defined in GO) has been widely investigated in recent years \cite{zhao2020literature,unsal2022learning}.
	A large part of these works such as GOLabler \cite{you2018golabeler} focus on exploring feature extraction, feature ensemble, and automatic feature learning of the proteins. For example, GOLabler \cite{you2018golabeler} utilizes five kinds of different protein sequence information while DeepGraphGO \cite{you2021deepgraphgo} builds a network of proteins and learns protein features via a Graph Neural Network. Recent methods attempt to further exploit inter-function (class) relationships that are defined in GO for better performance.
	For example, DeepGOZero \cite{kulmanov2022deepgozero} and HMI~\cite{xiong2022hyperbolic} use formal semantics including the class hierarchy, class disjointness axioms and complex class restrictions in OWL as additional constraints for training the multi-label classifier for protein function prediction. 
	Protein function prediction is a representative multi-label classification problem where complex relationships of the labels are defined in a KG and can be used for performance augmentation. 
	It is quite common in machine learning applications in the life sciences, such as the above mentioned automated clinical coding where the codes' semantics are modeled by the ICD ontology, and ecotoxicological effect prediction where the multiple affected species of a chemical to predict form a taxonomy.

	\medskip
	\noindent\textbf{Predictions for Healthcare using Ontologies with Clinical Data.}
	Digital Healthcare involves predictions using clinical data and ontologies, including diagnosis (\eg rare diseases) and procedure predictions (\eg ICU readmissions). A related concept is personalized medicine, which is achieved through the matching and fusion of knowledge from diverse sources, and plays a significant role in the prediction tasks. This often involves matching multiple ontologies \cite{silva2022matching}, integrating curated databases (\eg pharmacogenomics, molecules and proteins knowledge bases), mining knowledge from scientific literature \cite{Xia2018MiningDR} and person-centered clinical knowledge extracted from EHR or claim data, with distinguishing risk factors or cohorts' demographics (\eg age and gender), which could enhance predictions related to adverse effects \cite{10.1093/bib/bbx099} or rare diseases for which there are not enough labeled datasets \cite{Alsentzer2022}. For example, SHEPHERD \cite{Alsentzer2022} incorporates a multi-relational KG (extracted from PrimeKG \cite{Chandak2022_primeKG}) of diseases, phenotypes and genes, and leverages patient simulated data to discover novel connections between patients' clinical, phenotype and gene information to accelerate the diagnoses of rare diseases. Knowledge-guided learning is achieved by training a GNN to represent each patient's subgraphs of phenotypes in relation to other gene, phenotype, and disease associations within the KG, such that embeddings are informed by all of the existing biomedical knowledge captured in the network topology.
	
	The approach in \cite{carvalho2023knowledge} constructs a KG (using expressive OWL ontologies) to predict ICU (intensive care units) readmission risk by enriching EHR data with semantic annotations from various biomedical ontologies in BioPortal. These predictions are based on KG embedding, such as RDF2vec, OPA2vec, and TransE, and classical machine learning methods, such as Logistic Regression, Random Forest, Naive Bayes and Support Vector Machines. Drawing from the Health \& Social Person-centric Ontology (HSPO)~\cite{HSPO}, which focuses on multiple clinical, social and demographic facets for a patient or cohort, the approach presented in \cite{theodoropoulos-2023} builds a person-centric KG (expressive OWL ontology with TBox and ABox) from structured and unstructured data in EHR). Subsequently, a representation learning approach using GNNs is used to predict readmissions to the ICU.

	\subsection{What are the challenges?}
	
	We present four of the open challenges to unlock the full potential of methods to advance knowledge  discovery for the life sciences using KGs, based on the use cases above.
	
	\begin{itemize}
		
		\item \textbf{How to incorporate the semantics from a KG in machine learning?} Many life science knowledge discovery tasks are modeled as a machine learning classification problem, whose input and output labels have additional valuable information in one or multiple external KGs. The challenge lies in extracting this information, optionally encoding it into vector representations, and injecting that knowledge into machine learning and pre-trained language models. Doing this effectively remains an important open challenge especially for protein-related pre-trained language models \cite{zhang2022ontoprotein, xu2023protst,zhou2023protein}. Besides improving the accuracy in knowledge discovery, injecting semantics from KGs can also contribute to making the model more explainable (see Section \ref{sec:KG_explainable_AI}), but to this end, much research is still required.

		\item \textbf{How to deal with the long-tail phenomenon in machine learning with KGs?} In machine learning classification for real-world life science knowledge discovery, the candidate labels often exhibit a long-tailed distribution, \ie a small ratio of them are common with a large number of training samples available, while most of them are infrequent or even have never appeared before. For example, imbalance in training data may occur for rare diseases or adverse drug effects that affect only a small portion of the population \cite{Alsentzer2022,huang2023zeroshot,dong2023raredisease}. KGs sometimes have encoded the relationships of the labels, and could be used to help train the model for predicting those long-tailed labels or enable the inference of such labels.

		\item \textbf{How to create an efficient multi-modal representation of knowledge to enable discovery?} Most current state-of-the-art methods build learned graph representations based on isolated modalities. Multimodal KGs can explicitly capture labeled nodes and edges, each with well-defined meanings, across heterogeneous node types, relations and modalities (such as text, images, protein sequences, molecules fingerprints, diseases and more) \cite{Chandak2022_primeKG,lam-otter}. Incorporating KGs with multiple modalities for representation learning requires computationally scalable methods to compute the initial embeddings for each modality, as a preliminary step to learn computable representations of large knowledge. Furthermore, robust learning techniques are needed for generalizing the learned representations to nodes with unseen or missing modalities, thereby enabling the discovery of new knowledge. An example would be inferring properties of proteins for which only the sequence is known.

		\item \textbf{How to efficiently utilize and fuse heterogeneous datasets, such as human-curated domain knowledge bases, scientific literature and person-centered health records, for knowledge discovery?} State of the art shows that representations can be enhanced by incorporating richer information available across different sources \cite{huang2023zeroshot,lam-otter, silva2022matching}. Bringing in more data during training is needed to learn representations that can be applied to a broader range of downstream prediction tasks. However, learning from large and diverse KGs requires addressing challenges such as alignment, noise handling, balancing rich expressive knowledge with scalability and dealing with knowledge inconsistency. Moreover, more robust learning methods are needed for generalizing the learned representation to multiple downstream tasks (\eg knowledge-aware transfer, zero-shot and few-shot learning \cite{chen2023zero}). An important aspect in this regard is addressing the disparity between all of the knowledge accessible during pre-training and the knowledge accessible or relevant for downstream fine-tuning \cite{huang2023zeroshot,lam-otter}.
		
	\end{itemize}
	
	\section{Knowledge Graphs for Explainable AI}\label{sec:KG_explainable_AI}
	
	Machine Learning (ML) and Artificial Intelligence (AI) methods are widely employed to tackle complex problems in many domains, including life sciences such as chemistry or biomedicine.
	Yet many of those methods operate as a ``black-box'', not enabling domain experts to understand the reasoning behind their predictions~\cite{karim2023explainable}.
	This is a major concern, especially for applications in areas with a potential impact on human lives, or areas with legally enforced accountability or transparency~\cite{rajabiK22}. Moreover, understanding the workings of AI methods is also crucial in the context of scientific applications, such as those described in Section \ref{sec:Knowledge-Discovery}, where explaining the prediction process can help elucidate natural phenomena \cite{duran2021dissecting}. 
	
	One way to address this issue is to employ the methods of eXplainable Artificial Intelligence (XAI). Although this is a topic long explored in the AI research community, there is still no widely-accepted definition of explainability, with many terms being used interchangeably, such as interpretability, comprehensibility, understandability and transparency~\cite{BARREDOARRIETA202082}.
	Barredo \textit{et al.} define explainability as the ability of a model to make its functioning clearer to an audience~\cite{BARREDOARRIETA202082}.
	A slightly different definition is given in the previous survey~\cite{Guidotti2018}: ``an interface between humans and a decision maker that is at the same time both an accurate proxy of the decision maker and comprehensible to humans''.
	Both definitions focus on the audience, for \textit{whom} is the model explainable, but the second suggests an explanation is another artifact produced by a model or alongside the model. 
	
	There are two distinguishable audiences in the context of the life sciences: scientists (researchers) and healthcare practitioners~\cite{tiddiS22}.
	For the first group, the explanation is used as a guide to understanding within life sciences research for scientific discovery.
	As a result, the explanation may exist in a well-bounded context of a hypothesis or research project.
	On the other hand, practitioners are involved directly in decisions with impact on healthcare.
	They need to consider the output of the model in an open context, and sometimes also to explain the output to a patient who is not a domain~expert.
	
	A number of approaches for XAI emerge from the literature and broadly contain two parts: (1) transparent box design, which includes algorithms such as decision trees, where models can be directly interpreted by users and therefore an explanation of an output results in simply following the decision paths that relate input to output; (2) post hoc interpretability, which provides an explanation to a black-box model using additional methods such as probing, perturbing, or by constructing surrogate models for general ML or AI methods~\cite{karim2023explainable,tiddiS22}.

	Utilization of KGs can greatly enhance XAI qualities as KGs are ideal for improving the model's interpretability, explainability, and understandability.
	Some methods are directly built around KGs and thus take full advantage of them.
	Examples of those methods may include methods that are using paths~\cite{10.1007/978-3-031-33455-9_1}, predicting links, or performing reasoning~\cite{Donadello2019PersuasiveEO}.
	Other methods can be enhanced using the KG (\eg \cite{DBLP:journals/corr/abs-2210-15985}).
	Yet the enhancement effect greatly depends on the place where KGs are employed and iteratively applied: \textit{pre-model} (\eg KG construction, potentially multi-modal), \textit{in-model} (\eg integrating KG with machine learning models), and \textit{post-model} (\eg reviewing and updating KG by domain experts to be applied in the next iteration to enhance machine learning models and their explanability)~\cite{rajabiK22}.
	For example in in-model use, a model can be pre-trained using a KG, and an example of a pre-trained language model is SapBERT \cite{liu-etal-2021-self}, which utilizes synonyms in the UMLS Metathesaurus to further pre-train a BERT language model. 
	This can not only be beneficial for performance~\cite{zhang2022ontoprotein}, but can also potentially enhance post-model explanation since the trained features are aligned with~the KG~\cite{rajabiK22}.

	\subsection{What has been done: use cases and recent developments}
	
	\noindent\textbf{Explainable AI for Healthcare Practice.}
	The utilization of AI in healthcare practice raises the concern of leaving life-critical decisions to black-box models~\cite{rajabiK22,tiddiS22}.
	For example, in the field of precision medicine which aims at tailoring drug treatments and dosages to each patient, clinicians require more information from a model than a simple binary decision~\cite{BARREDOARRIETA202082}.
	The interpretability and explainability of AI models is thus an essential characteristic to make outputs understandable and transparent.
	This would enforce both clinicians' and patients' trust in models by complementing (and not substituting) clinicians' explanations~\cite{chariSGFDM20,rajabiK22,tiddiS22}.
	
	To illustrate, this direction has been envisioned for several healthcare scenarios.
	Explainable AI models could support the experts in finding clinical trials that are appropriate based on patient history~\cite{tiddiS22}.
	Counterintuitive or unreliable predictions that could have serious consequences could be explained, and thus prevented~\cite{tiddiS22,caruanaLGKSE15,karim2023explainable}.
	Some also envision such models to be used to explain and debunk healthcare-related misinformation~\cite{rajabiK22}.
	As aforementioned, it is noteworthy that different kinds of explanations should be employed depending on the target audience, \eg scientific explanations for evidence or trace-based explanations for treatment~\cite{chariSGFDM20}.
	
	\medskip
	\noindent\textbf{Explainable AI for Knowledge Discovery.}
	As introduced in Section~\ref{sec:Knowledge-Discovery}, KGs can support knowledge discovery in life science, including 
	the explainability of the process and the discovered units.
	In this view, Ritoski and Paulheim~\cite{ristoskiP16} explain that ontologies, linked data, and KGs are used in the interpretation step of a data mining process, \eg  for interpreting sequential patterns in patient data~\cite{jayd13}, or to describe subgroups in a semantic subgroup discovery process~\cite{trajkovskiLT08}.
	KGs can also serve both as the basis for knowledge discovery processes and the interpretation process.
	For example, Linked Open Data connecting drugs and adverse reactions can be analyzed with Hidden Conditional Random Fields to predict adverse drug reactions, where the paths from selected drugs to outcomes visually explain the prediction~\cite{kamdarM17}.
	Similarly, Bresso \textit{et al.}~\cite{bressoMBCNPSC21} leverage features extracted from KGs (interpretable features such as paths, neighbors, path patterns) and white box models (\eg decision trees) to reproduce expert classifications of drugs causing or not specific adverse drug reactions. 
	The rules extracted from the decision trees contain features that provide explanations for the molecular mechanisms behind these adverse reactions according~to~experts. Sousa \textit{et al.}~\cite{sousa2023explainable} employ KGs to explain both protein-protein interaction predictions and gene-disease association predictions based on shared semantic aspects.
	
	\medskip
	\noindent\textbf{Explainable AI for KG Construction}
	The final use case considers the situation that XAI is applied to KGs themselves. We discussed the challenge to support human intervention in KG construction in Section \ref{sec:kgc-challenges}. Recent KG construction gradually relies on data-driven, deep learning based methods to automatically induce knowledge from data. The deep learning models are opaque, and thus the process requires explainability. The resulting KG may not be accountable to be used for downstream applications. \textit{Trustworthy KG engineering} is proposed in \cite{zhang2023expHumanKGC} to highlight the importance of embedding explainable AI and human intervention in the KG life cycle. XAI methods have been applied in many NLP related tasks (entity and relation extraction, entity resolution, link prediction, etc.) in KG construction from texts. The XAI methods rely either on feature-based explanations or knowledge-based explanations. While feature-based explanations try to infer explanations from the data or the models' interpretation of the data, knowledge-based explanations aim to interpret the process with rules, reasoning paths, and structured contextual information. Rules and paths have mainly been used for explanation, especially for link prediction, a task comprehensively surveyed in \cite{zhang2023expHumanKGC}. 
	
	\subsection{What are the challenges?}
	
	\begin{itemize}
		
		\item \textbf{How to integrate KGs for better XAI, especially with recent deep learning and language model based methods?} KG may provide better data provenance for the model output. This can ensure explainability for communicating the model to domain experts in data science applications~\cite{BARREDOARRIETA202082}. In terms of recent generative LLMs, life science KGs, with careful curation based on scientific publications, may help to provide provenance data to the answers generated by LLMs. Studies need to understand to what extent, and how, LLMs can be applied to induce knowledge (\eg by probing LLMs with biomedical ontologies \cite{he2023language}), which then may provide a foundation to create better approaches to integrate KGs with LLMs. Another area is neuro-symbolic methods which may provide models that are inherently more interpretable (see further discussions in Section \ref{sec:overall}). Also, regarding language models (especially LLMs), they are capable of generating fluent texts, which can potentially serve as textual explanation generators from symbolic knowledge for XAI. Meanwhile, a key issue is the hallucination of LLMs, and KGs may support better prompting, fine-tuning and interpretable inference of LLMs for higher decisiveness and trustworthiness   \cite{pan2023unifyingLLMandKG}. 
		
		\item \textbf{How to evaluate XAI methods that involve KG?} How to measure the quality of explanations, to ensure they are corresponding to users? The majority (around 70\%) of XAI studies for KG construction do not evaluate the quality of the explanations or only informally visualize or comment on a limited number of cases to show the intuitive outcome \cite{zhang2023expHumanKGC}. Also, an XAI method needs to consider the target audience, as the explainability is to be finally received by a group of humans~\cite{BARREDOARRIETA202082}. For instance,  only a small number of current approaches to XAI for KG construction involve a user study, human evaluation or task-specific metrics~\cite{zhang2023expHumanKGC}.
		Evaluating the quality of explanations requires some expert evaluation performed as ex-post evaluation, and well-defined metrics are needed for this task. An example is in~\cite{halliwellGL21} to use a combination of users' scores for each predicted explanation in a KG link prediction task, where there are multiple possible explanations. More expert validated and automated evaluation methods and associated metrics are required for KG-related XAI.
		
	\end{itemize}
	
	
	\section{Discussion and Conclusion}\label{sec:Discussion}
	
	In this work, we have summarized the recent developments of KG research in life science on three important topics -- KG Construction and Management, Life Science Knowledge Discovery, and KG for XAI. While each topic has its specific challenges, there are some common challenges and trends for the life science KG research in general.
	
	\subsection{Overall challenges and trends}
	\label{sec:overall}
	
	Meanwhile, more scalable and efficient knowledge retrieval, query and reasoning systems, including life science KGs and mapping repositories, are still worthy of investigation and development.
	
	\medskip
	\noindent\textbf{Evolution and Quality Assurance of KGs.} KGs need to be updated as new data and knowledge are emerging, and the schema and facts can easily become outdated or less useful for existing applications in life sciences. In terms of KG construction, we discussed ontology extension as a use case to address the evolution issue or emergence of new concepts and relations, and also instance matching to extend new instances for the KG. Updating KGs is also a prerequisite for life science knowledge discovery and knowledge discovery methods should be able to support the evolution of KGs with \eg the capabilities of continuous learning and zero-shot learning. 
	Quality assurance is another issue for KGs, including the tasks of knowledge error detection and correction, knowledge completion, knowledge canonicalization, etc. 
	On the one hand, more effective KG quality assurance methods and systems should be developed, including schema and constraint languages for quality verification and learning-based models for prediction (\eg \cite{chen2023assertion} combines both for fact correction); on the other hand, knowledge discovery methods should be robust to noisy KGs by investigating \eg robust KG embeddings and multi-modal representation~learning.

	\medskip
	\noindent\textbf{Heterogeneity in KGs: Multi-domain and Multi-modality.}
	KGs contain heterogeneous information, which brings challenges to their construction, representation, and reasoning. Different schema and data in KGs can have different focuses in their scopes and domains. Integrating data of different domains for building \textit{multi-domain} KGs is difficult with challenges in \eg ontology and data matching.
	Besides, recent studies have explored integrating different modalities to construct \textit{Multi-modal} KGs \cite{chen2023survey,MURALI2023104403,wang2023cross}, for instance text \cite{pahuja2021systematic}, images \cite{wang2022wikidiverse}, etc. One challenge to address is how to learn effective machine learning models over multi-modal KGs fused from different sources (patients' records, curated knowledge bases, and scientific literature) to support scientific discovery as well as KG construction and management.
	Another challenge is developing accurate and efficient knowledge representation approaches for texts and images in multi-modal KG construction. For example, careful consideration should be given to when to simply use an annotation property to associate an image with an entity, and when to use a property with specific semantics to connect an image and an entity.
	
	\medskip
	\noindent\textbf{Human Interaction and Explainability with KGs.} In KG construction, human experts are required for many sub-tasks of KG construction and provide oversight \cite{zhang2023expHumanKGC}. In life science knowledge discovery, human experts are necessary to finally validate the predicted new knowledge. The whole process of interacting with KG in life sciences requires explainability, especially when sub-symbolic models (\eg pre-trained language models) are used. How to generate clear explanations for human interaction and how to evaluate the quality of explanations remains a challenge, as well as how to achieve consensus regarding scientific understanding with automatically discovered knowledge when organizing knowledge in life science \cite{neuhaus2022ontology}. The recent growth of \textit{Neuro-Symbolic methods} suggests their support for explainability \cite{Karim2022phdthesis,karim2023explainable,rodriguezLSFDTF22}. A recent survey  \cite{karim2023explainable} summarizes XAI in bioinformatics with a chapter on knowledge-based explanations, whereas Karim \cite[Chapter 8]{Karim2022phdthesis} provides a neuro-symbolic framework for KG construction and utilization for medical experts' decision making in the cancer domain. The approach presented in \cite{rodriguezLSFDTF22} is another recent example of neuro-symbolic integration for image classification with KG-based XAI in the cultural~heritage~domain.
	
	\medskip
	\noindent\textbf{Personalized and Customized KGs}. A key challenge for KG construction is customization, as we discussed in Section \ref{sec:KG_construc}, to construct application-oriented KGs, where relevant sub-KGs have to be extracted for large-scale KGs (\textit{a.k.a.} modularization) and integrated with other knowledge and data from different sources. Besides, many life science KGs are about individuals, \eg patients in healthcare applications, where Personal Health KG enables the integration of instance-level (or patient-level) information and their computation is required \cite{MURALI2023104403}. An example is the Personal Health KG in \cite{chen2022phkg} that supports the dietary recommendation for users, where the construction and population of the KG requires reusing and integrating existing ontologies, dietary guidelines, and time-series patient data. The other examples of KGs integrating patients' EHR data \cite{theodoropoulos-2023,carvalho2023knowledge} are presented in Section \ref{sec:KnowDisc-dev}. In personal KG construction, personal data should be protected. KG scalability should also be considered in order to be used on small devices such as cellphones. This is still a big challenge that has been rarely considered in using KGs in the life sciences.
	
	\medskip
	\noindent\textbf{Distributed KGs.}
	The value of healthcare data for improving clinical knowledge and standard of care and the potential of semantic technologies to further enhance it are well recognized. However, a responsible use of healthcare data at the global level (beyond each healthcare provider and even each country) must take into account both legal and ethical issues in data sharing, privacy and security.  Distributed knowledge graphs can mitigate these issues, by allowing for access control and privacy protection. Furthermore, distributed knowledge graphs can also address the challenges of scientific data ownership and stewardship by enabling the decentralized publishing of high quality data. Several approaches for federated querying and embedding of knowledge graphs have been proposed in recent years~\cite{chen2022federated, peng2021differentially, sima2019enabling}, however a wide adoption of semantic technologies in healthcare is still lacking, with a proliferation of terminological standards and a disconnection between data and meaning.

	\medskip
	\noindent\textbf{Representation Learning with KGs: Symbolic and Sub-symbolic Integration}. Across the topics and use cases, we see the importance of transforming symbolic knowledge into sub-symbolic representations or combining both representations. The combination of both the neural and the traditional symbolic representation methods leads to a trend in neural-symbolic approaches in the field \cite{acm-ml-sw-2023}. Recently, Pre-trained and Large Language Models provide new methods to transfer self-supervised learning from a vast amount of corpora to support KG construction, \eg OntoGPT \cite{caufield_structured_2023} and OntoLAMA \cite{he2023language}. LLMs are especially good at representing texts of life science publications in sub-symbolic spaces for semantic understanding.
	KGs may also provide a layer of explainability by validating the output of LLMs. A recent survey \cite{pan2023unifyingLLMandKG} proposes a roadmap for integrating LLMs and KGs.  
	OntoProtein \cite{zhang2022ontoprotein} is a recent example of how to integrate KGs into the process of pre-training LLMs in the bioinformatic domain, thus achieving improved results on protein-related knowledge discovery tasks.
	Also, geometry-informed representations of more formal KGs, especially in hyperbolic spaces or using complex geometric structures, \eg \cite{chami2020low,kulmanov2022deepgozero}, can usually represent the structure of the KG with low dimensional vectors. Graph Neural Networks may also support the encoding of KG structures in a more explainable way with logical rules \cite{cucala2021explainable}.

	\subsection{Conclusion}
	Knowledge Graphs have become a popular and effective method to represent heterogeneous concepts, relations, and data in life sciences. They require scalable solutions to represent and reason with heterogeneous data and require constant updates. Throughout this work, we covered the main topics and their corresponding use cases of KGs in multiple life science domains such as protein analysis, drug discovery, ecotoxicology, and healthcare, and summarized the corresponding challenges. As new methods in knowledge representation appear, for instance the recent trends of human-in-the-loop, sub-symbolic knowledge representations, pre-trained and large language models, and neuro-symbolic integration, we envisage deeper applications of KGs to life science processes, that support the construction of  more applicable KGs and the discovery of more reliable scientific knowledge, with explainability and human interaction better supported.
	KGs in combination with other modern machine learning and natural language processing techniques will become a foundation for AI for the life sciences. 
	
	\paragraph*{Appendix A: Terms in Knowledge Graphs and Life Sciences}\label{special-key-terms}
	
	Below we provide a list of key terms used in this paper, as well as their definitions and explanations. Note we mainly use the original sentences in the sources that are referenced as the definitions.
	
	\textbf{Description Logics}: a family of knowledge representation languages that can be used to represent knowledge of an application domain. DLs differ from their predecessors, such as semantic networks and frames, in that they are equipped with logic-based semantics, the same semantics as that of classical first-order logic. Most ontologies are implemented in OWL, whose semantics are given by the Description Logic $\mathcal{SROIQ}$. \cite{baader_horrocks_lutz_sattler_2017}
	
	\textbf{TBox} and \textbf{ABox}: the two components of domain knowledge in Description Logics, \ie a terminological part called the TBox and an assertional part called the ABox, with the combination of a TBox and an ABox being called a knowledge base (KB). The TBox represents knowledge about the structure of the domain (similar to a database schema), while the ABox represents knowledge about a concrete situation (similar to a database instance). \cite{baader_horrocks_lutz_sattler_2017}
	
	\textbf{Semantic Networks}: a graph structure for representing knowledge in patterns of interconnected nodes and arcs \cite{sowa1992semantic}. We use the term to denote a graph of concepts and relations without formal semantics.
	
	\textbf{Gene Ontology}: The Gene Ontology (GO) knowledgebase provides a comprehensive, structured, computer-accessible representation of gene function, for genes from any cellular organism or virus \cite{ashburner2000gene,GO2023}.
	
	\textbf{SNOMED-CT}: Systematized Nomenclature of Medicine Clinical Terms (SNOMED CT) is a structured clinical vocabulary. It has a general and comprehensive coverage of clinical terms to support electronic healthcare systems and clinical applications. \cite{donnelly2006snomed,chap23HIbookCoiera}
	
	\textbf{UMLS} (\textbf{UMLS Metahesaurus} and \textbf{UMLS Semantic Networks}): Unified Medical Language System (UMLS) is a repository of biomedical vocabularies developed by the US National Library
	of Medicine. The UMLS is composed of three ``knowledge sources'', a Metathesaurus, a semantic network, and a lexicon. The UMLS Metathesaurus is a comprehensive effort for integrating biomedical
	ontologies through mappings. The UMLS Semantic Networks define the types or categories, or Semantic Types, of all Metathesaurus concepts and their relationships, or Semantic Relations. \cite{DBLP:journals/nar/Bodenreider04,chap23HIbookCoiera}
	
	\textbf{ChEBI}: Chemical Entities of Biological Interest (ChEBI) is a database and ontology containing information
	about chemical entities of biological interest. \cite{hastings2016chebi}
	
	\textbf{Symbolic vs. subsymbolic representations}: Rooted in cognitive science, symbolic systems of human cognition are related to the representation and manipulation of symbols; sub-symbolic or connectionist systems are most generally associated with the metaphor of a neuron, \eg perceptrons as an early system \cite{Kelley2003symb-and-subsymb}. In terms of AI, symbolic systems contain logic-based and knowledge representations, while subsymbolic systems typically contain neural networks and deep learning based methods \cite{Garcez2009neural-sym}. Neural language models and pre-trained language models \cite{jurafsky2023slpbook} are also classified under subsymbolic systems.
	
	\textbf{Pre-trained and Large Language Models}: Neural language modeling is the task of using neural network approaches to predict words from prior their contexts in a sequence. Pre-training is the process of learning some sort of representation (usually neural embedding based) of meaning for words
	or sentences by processing very large amounts of text (or other data in a sequence form, \eg proteins and KG facts). This results in pre-trained language models. The dominating architecture for neural language modeling is Transformer-based models, including BERT, its domain specific versions, and later large variants, like the GPT series. The pre-trained language models of very large sizes are recently coined Large Language Models (LLMs). \cite{jurafsky2023slpbook}
	
	\textbf{Neuro-symbolic representations}: refers to the integration of neural networks and symbolic representations to design AI models that base their prediction on both data and knowledge. \cite{Garcez2009neural-sym}
	
	\paragraph*{Appendix B: Authors' Contributions}
	All authors participated in the planning and discussions of this work. JH and HD finished the abstract and ``Introduction''. VT, JC and EJR contributed to ``Knowledge Graphs in the Life Sciences''. VT contributed to the main part of ``Knowledge Graph Construction and Management'', with contributions of use cases from JC, HD, PM, EJR, and JH. VL and JC contributed to ``Life Science Knowledge Discovery''. PM, PS, HD, and CP contributed to ``Knowledge Graphs for Explainable AI''. HD, JC, and CP contributed to ``Discussion and Conclusion'' based on discussions with other team members. All authors contributed to the final revision of this paper.
	
	\bibliography{ref}
	\bibliographystyle{plainurl}
	
\end{document}